\documentclass[runningheads]{llncs}
\usepackage[T1]{fontenc}
\usepackage{graphicx}
\usepackage{amsmath,amssymb,amsfonts}
\usepackage{xcolor}
\usepackage{colortbl}
\usepackage[export]{adjustbox}
\usepackage{arydshln} 
\usepackage{cite}

\begin{document}

%
\title{Promoting Shape Bias in CNNs: Frequency-Based and Contrastive Regularization for Corruption Robustness}
%
%
\author{Robin Narsingh Ranabhat \and
Longwei Wang \and
Amit Kumar Patel \and
KC santosh 
}

\institute{USD AI Research Lab, University of South Dakota, Vermillion SD 57069, USA 
\\
\email{Emails: robin.ranabhat@coyotes.usd.edu, longwei.wang@usd.edu, amit.patel@coyotes.usd.edu, kc.santosh@usd.edu}}

\maketitle              

\begin{abstract}
Convolutional Neural Networks (CNNs) excel at image classification but remain vulnerable to common corruptions that humans handle with ease. A key reason for this fragility is their reliance on local texture cues rather than global object shapes—a stark contrast to human perception. To address this, we propose two complementary regularization strategies designed to encourage shape-biased representations and enhance robustness. The first introduces an auxiliary loss that enforces feature consistency between original and low-frequency filtered inputs, discouraging dependence on high-frequency textures. The second incorporates supervised contrastive learning to structure the feature space around class-consistent, shape-relevant representations. Evaluated on the CIFAR-10-C benchmark, both methods improve corruption robustness without degrading clean accuracy. Our results suggest that loss-level regularization can effectively steer CNNs toward more shape-aware, resilient representations.

\end{abstract}

\section{Introduction}
Convolutional Neural Networks have achieved remarkable success on benchmark datasets such as ImageNet and CIFAR-10, demonstrating high accuracy in a wide range of image classification tasks. However, their performance significantly degrades in the presence of common image corruptions, such as noise, blur, compression artifacts, and environmental distortions like fog or frost \cite{b1,wang1,wang2,wang3,wang4,wang5,wang6,wang7,wang8,wang9,wang10}. This brittleness undermines the practical deployment of CNNs in real-world applications, including autonomous driving, medical imaging, and surveillance systems, where input data is rarely clean or curated. 
Recent work by Geirhos et al.~\cite{b2} revealed that standard CNNs trained on datasets like ImageNet predominantly rely on local texture patterns rather than global object shape for making predictions. This observation contrasts sharply with the human visual system, which is known to prioritize shape cues for object recognition. The reliance on texture renders CNNs highly susceptible to corruptions that primarily perturb high-frequency components—textures—while leaving shape information relatively intact.

\begin{figure}[t]
\centering
\begin{minipage}[t]{0.48\textwidth}
  \centering
  \includegraphics[width=\linewidth]{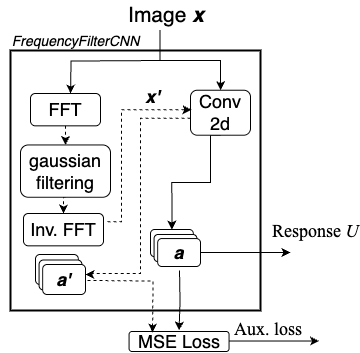}
  \caption{\small Custom CNN layer which produces a separate reconstruction of original input \textit{x} based on low-frequency components obtained after Fourier Transform. Activations \textit{a} and \textit{a'} are generated separately from the same Convolution layer. \textit{a'} is cached to calculate the Mean-Squared-Loss.}
  \label{fig:fourier-reconstruction}
\end{minipage}
\hfill
\begin{minipage}[t]{0.48\textwidth}
  \centering
  \includegraphics[width=\linewidth]{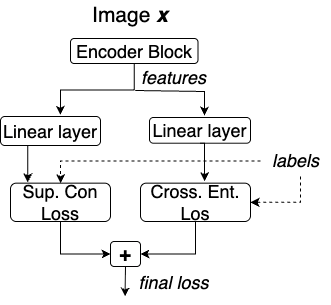}
  \caption{\small Integrating Supervised Contrastive Loss with Regular Cross-Entropy Loss. We hypothesize this enables learning high-level shape features existent within images of same class.}
  \label{fig:constrastive-loss}
\end{minipage}
\end{figure}

Motivated by this perspective, we propose two orthogonal yet complementary regularization strategies that modify the loss function to encourage CNNs to learn representations that may exhibit reduced texture bias and enhanced shape focus. Our approach centers on the principle that appropriate regularization can guide the optimization process toward solutions that generalize better under distribution shift, particularly corruption-induced changes.
Our first regularization approach operates in the frequency domain by introducing auxiliary loss terms that enforce consistency between feature activations of original images and their low-frequency filtered counterparts. This regularization mechanism biases the network toward learning features that remain stable across frequency manipulations, potentially favoring shape-relevant information that typically resides in low-frequency components over texture-specific high-frequency details.
Our second regularization strategy employs supervised contrastive learning as an additional term in the training objective. Unlike traditional cross-entropy training that focuses solely on output-level classification accuracy, this regularization approach explicitly structures the intermediate feature space by pulling together representations of same-class samples while pushing apart different-class representations. We hypothesize that this representation-level regularization naturally encourages the learning of class-invariant features, which are more likely to be shape-based than texture-based. The key insight is that strategic regularization can indirectly promote beneficial inductive biases without requiring explicit architectural changes or specialized data preprocessing. 
We evaluate our regularization methods on the CIFAR-10-C benchmark, which applies 19 diverse corruption types to the CIFAR-10 dataset across five severity levels, providing a rigorous testbed for corruption robustness. Our results demonstrate that both regularization strategies improve robustness while maintaining clean accuracy, showing complementary improvements across different corruption types.




\section{Related Work}\label{sec:related}
\subsection{Robustness and Regularization in CNNs}
 CNNs struggle under distributional shifts caused by common image corruptions. To assess this vulnerability, Hendrycks and Dietterich~\cite{b1} introduced the ImageNet-C and CIFAR-10-C benchmarks, applying real-world corruptions like Gaussian noise and motion blur. Their results showed significant performance drops even in state-of-the-art models.

In response, numerous regularization strategies have been proposed to enhance model robustness. Classical regularization techniques like dropout~\cite{b2} and batch normalization~\cite{b3} have been shown to improve generalization, but their specific impact on corruption robustness has been less systematically studied. Data augmentation remains a widely used approach, wherein training data is augmented with synthetic corruptions to increase diversity and reduce overfitting. Notably, methods such as AugMix~\cite{b4} have shown promise in this regard. However, these techniques often require careful design and may struggle to generalize across all corruption types. Alternative solutions include neuroscience-inspired architectural enhancements, such as divisive normalization layers~\cite{b5} and push-pull inhibition mechanisms~\cite{b6,b7,wang11,wang12,wang13,wang14,wang15,wang16,wang17,wang18}, which aim to emulate biological visual processing. 

\subsection{Shape vs. Texture Bias}
Geirhos et al.~\cite{b8} demonstrated that conventional CNNs are strongly biased toward texture-based features rather than shape-based cues. This conclusion was drawn using Stylized ImageNet (SIN), a modified version of ImageNet in which textures were randomized via style transfer while preserving object shapes. CNNs trained on SIN showed increased shape bias and improved robustness to corruptions. This observation stands in contrast to human visual perception, which primarily relies on global shape for object recognition. The gap between CNNs and human perception has motivated follow-up studies exploring techniques to enhance shape bias in models. Some approaches include designing networks with greater receptive fields to better capture global structures~\cite{b9,li2019dft}, implementing advanced data augmentation techniques that preserve structural information~\cite{b10} and adopting vision transformer architectures, which inherently prioritize global shape due to their patch-based attention mechanisms~\cite{b11}.

\subsection{Fourier Perspective on Robustness}
Another lens through which CNN robustness has been analyzed is the frequency domain. Yin et al.~\cite{b12} showed that CNNs tend to rely heavily on high-frequency components, which are particularly sensitive to noise and artifacts. This reliance renders models fragile under conditions that perturb fine-grained details. Building on this insight, Wang et al.~\cite{b13} proposed training strategies that encourage models to focus on low-frequency components—typically associated with shape—thereby improving generalization and corruption robustness. These findings support the hypothesis that promoting low-frequency, shape-relevant representations can serve as a foundation for robust visual learning.

\subsection{Contrastive Learning}
Contrastive learning has recently gained prominence as a powerful framework for representation learning, particularly in the context of self-supervised learning~\cite{b14}. It operates by structuring the latent space such that similar samples (positive pairs) are drawn closer together while dissimilar ones (negative pairs) are pushed apart. This principle has been extended to supervised learning by Khosla et al.~\cite{b15}, who demonstrated that incorporating label information into the contrastive objective significantly improves the quality of learned representations compared to standard cross-entropy training.
While contrastive learning has predominantly been applied to improve general feature quality and transferability, its potential to enhance model robustness—particularly through promoting shape-based invariance—remains underexplored. Our work builds upon this emerging direction by employing supervised contrastive loss to foster representations that are both discriminative and stable across corruptions, with an emphasis on capturing class-specific shape features.

\section{Proposed Methodology}\label{sec:method}
We propose two complementary regularization approaches to improve CNN robustness to common corruptions by modifying the training objective through auxiliary loss terms. 


\subsection{Frequency-Based Regularization with Filtered Convolution}
Our first approach is motivated by the observation that texture information typically resides in high-frequency components of images, while shape information is predominantly captured by low-frequency components. We introduce a novel convolution layer, \textit{FrequencyFilterCNN}, that encourages the network to produce similar activations for both original inputs and their high-frequency filtered counterpart as shown in Fig.~\ref{fig:fourier-reconstruction}. The \textit{FrequencyFilterCNN} operates as follows:
\begin{enumerate}
\item For an input tensor $x$, compute standard convolution activations $a_x = Conv(x)$.
\item Transform $x$ into the frequency domain using the 2D Fast Fourier Transform (FFT): $X = \mathcal{F}(x)$.

\item Apply a Gaussian low-pass filter in the frequency domain to obtain $X'$, suppressing high-frequency components: $G(u,v) = exp{(-{(u-u_c)^2 - (v-v_c)^2}/{2\sigma^2})}$, where $(u_c, v_c)$ is the center of the frequency domain, and $\sigma$ controls the cutoff frequency.

\item Transform back (inverse FFT) to the spatial domain: $x' = \mathcal{F}^{-1}(X')$. $x'$ also represents a high-frequency filtered version of original input $x$. 

\item Compute activations for the filtered input: $a_{x'} = Conv(x')$.

\item During training, minimize the mean squared error between $a_x$ and $a_{x'}$ as an auxiliary loss: $\mathcal{L}_{\text{aux}} = \text{MSE}(a_x, a_{x'})$
\end{enumerate}
The final loss function combines the standard cross-entropy loss with this auxiliary loss: $\mathcal{L}_{\text{total}} = \mathcal{L}_{\text{CE}} + \lambda \cdot \mathcal{L}_{\text{aux}}$, where $\lambda$ is a hyperparameter controlling the contribution of the auxiliary loss.
By encouraging similar activations for both original and high-frequency filtered inputs, the network is biased toward features that are preserved under blurring, which we hypothesize to be more shape-related rather than texture-related. This approach differs from simply training over additional blurred images, as we have control during backpropagation to update weights of specified CNN layers. Ideally, we only swap the first few CNN layers with \textit{FrequencyFilterCNN}, while using standard CNN for rest of downstream layers, assuming, the top layers will extract discriminative features from noisy inputs ~\cite{b6}~\cite{b7}.

\subsection{Supervised Contrastive Learning}
Our second approach leverages supervised contrastive learning to encourage the network to learn representations that capture class-invariant features, which are more likely to be shape-related than texture-related. 
As visualized in Fig.~\ref{fig:constrastive-loss}, we modify a standard ResNet architecture to produce two representations from the penultimate layer: $f(x)$ used for contrastive learning and $g(x)$ used for classification with cross-entropy loss. We adopt the supervised contrastive loss from Khosla et al. \cite{b15}:
\begin{equation}
\mathcal{L}_{\text{SupCon}} = \sum{i \in I} \frac{-1}{|P(i)|} \sum_{p \in P(i)} \log \frac{\exp(f_i \cdot f_p / \tau)}{\sum_{a \in A(i)} \exp(f_i \cdot f_a / \tau)}
\end{equation}
where $I$ is the set of indices in the batch, $P(i)$ is the set of indices of examples with the same class as example $i$, $A(i) = I \setminus {i}$, and $\tau$ is a temperature parameter. The representations $f$ are normalized to the unit hypersphere.
Similar to earlier, the final cross-entropy loss is regularized using this supervised contrastive loss as: $\mathcal{L}_{\text{total}} = \mathcal{L}_{\text{CE}} + \alpha \cdot \mathcal{L}_{\text{SupCon}}$, where $\alpha$ is a hyperparameter controlling the contribution of the contrastive loss. Since shape is more consistent within classes than texture (which can vary considerably even within different instances of same class), we consider this approach to naturally promote the shape bias.

\section{Experimental Details}\label{sec:implementation}
\subsection{Datasets and Model Design}
We conducted our experiments using the CIFAR-10 dataset \cite{b16} for training and the CIFAR-10-C benchmark \cite{b1} for evaluating robustness. CIFAR-10 consists of 50,000 training images and 10,000 test images across 10 classes, with each image being 32×32 pixels. CIFAR-10-C applies 19 different corruption types at 5 severity levels to the CIFAR-10 test set, resulting in 95 test scenarios (19 corruptions × 5 levels).
We applied minimal data augmentation during training, consisting only of random cropping and horizontal flipping. For both methods, we used ResNet-18 ~\cite{b17} as our backbone architecture, with following modifications specific to each method:

\subsubsection{Low-Frequency Activation Learning}
We replaced the first three convolutional layers of ResNet-18 with our custom \textit{FrequencyFilterCNN}. The \textit{FrequencyFilterCNN} was implemented as described in Section 3.1, with the Gaussian filter's sigma parameter set to $0.1$.

During training, we computed the auxiliary MSE loss across for each \textit{FrequencyFilterCNN} module in the network between their $a_x$ and $a_{x'}$ activations and summed them to get final MSE loss. This loss was scaled by a factor $\lambda = 0.2$ and added to the cross-entropy loss.

\subsubsection{Supervised Contrastive Learning}
For the supervised contrastive learning approach, we kept the standard ResNet-18 architecture but added two separate linear heads after the final global average pooling layer. First is a projection head for contrastive learning, consisting of a linear layer followed by L2 normalization. Second one is a classification head for standard classification, consisting of a single linear layer. The projection head mapped the flattened ResNet features to a 64-dimensional embedding space, where the contrastive loss was computed. The classification head mapped the encoded features to the probability distributions over the 10 output classes.

\begin{table}[t]
\caption{Accuracy (\%) of different models on clean CIFAR-10 and various corruption types from CIFAR-10-C. For readability purpose, the best results are in \textbf{bold} and the second best results are \underline{underlined}.}
\centering
\footnotesize
\renewcommand{\arraystretch}{1.0}
\setlength{\tabcolsep}{3pt}
\begin{tabular}{|>{\raggedright\arraybackslash}m{0.24\linewidth}
>{\centering\arraybackslash}m{0.18\linewidth}  
>{\centering\arraybackslash}m{0.18\linewidth}
>{\centering\arraybackslash}m{0.18\linewidth} 
>{\centering\arraybackslash}m{0.18\linewidth} 
|}
\hline
\textbf{Corruption Type}
& \textbf{ResNet-18 Baseline} &
\textbf{Frequency Domain Reg. (Ours)} & 
\textbf{Supervised Contrastive Reg. (Ours)} & 
\textbf{Push-Pull Baseline} \\
\hline \hline
\textbf{Clean CIFAR-10} & \underline{84.82} & 84.23 & \textbf{86.19} & 82.63 \\
\rowcolor{gray!10}
\textbf{Mean Corruption Acc} & 68.33 & 68.60 & \underline{69.13} & \textbf{70.31} \\
\textit{Gaussian Noise} & \underline{45.07} & 44.77 & 42.59 & \textbf{62.90} \\
\rowcolor{gray!10}
\textit{Impulse Noise} & 53.44 & 45.91 & \underline{54.58} & \textbf{62.92} \\
\textit{Shot Noise} & 52.90 & \underline{53.02} & 52.39 & \textbf{67.82} \\
\rowcolor{gray!10}
\textit{Speckle Noise} & 54.72 & 54.79 & \underline{54.97} & \textbf{68.33} \\
\textbf{Avg. Noise Acc.} & \underline{51.53} & 49.62 & 51.13 & \textbf{65.5} \\
\rowcolor{gray!10}
\textit{Gaussian Blur} & 69.32 & \textbf{71.50} & \underline{69.92} & 69.03 \\
\textit{Defocus Blur} & 74.32 & \textbf{75.92} & \underline{75.44} & 73.53 \\
\rowcolor{gray!10}
\textit{Glass Blur} & 71.30 & \textbf{74.02} & 71.32 & \underline{72.53} \\
\textit{Motion Blur} & 68.98 & \textbf{70.06} & \underline{69.84} & 68.61 \\
\rowcolor{gray!10}
\textit{Zoom Blur} & 70.52 & \textbf{72.60} & \underline{72.17} & 70.86 \\
\textbf{Avg. Blur Acc.} & 70.89 & \textbf{72.82} & \underline{71.74} & 70.91 \\
\rowcolor{gray!10}
\textit{Brightness} & \underline{80.98} & 80.90 & \textbf{82.93} & 78.89 \\
\textit{Contrast} & \underline{53.79} & 52.55 & \textbf{54.56} & 49.74 \\
\rowcolor{gray!10}
\textit{Elastic Transform} & 77.06 & \underline{77.63} & \textbf{77.93} & 75.40 \\
\textit{Fog} & \underline{70.78} & 68.84 & \textbf{71.37} & 64.88 \\
\rowcolor{gray!10}
\textit{Frost} & 71.45 & \textbf{74.58} & \underline{72.59} & 70.45 \\
\textit{JPEG Compression} & 74.66 & 75.23 & \underline{76.37} & \textbf{76.79} \\
\rowcolor{gray!10}
\textit{Pixelate} & \underline{81.58} & \textbf{82.44} & 81.45 & 80.45 \\
\textit{Saturate} & \underline{74.80} & 74.74 & \textbf{77.50} & 73.45 \\
\rowcolor{gray!10}
\textit{Snow} & 74.09 & \textbf{75.20} & \underline{75.13} & 72.25 \\
\textit{Spatter} & 78.60 & \underline{78.78} & \textbf{80.47} & 77.11 \\
\hline
\end{tabular}
\label{corruption_accuracy_table}
\end{table}

\begin{figure}[t]
    \centering
    \includegraphics[width=\textwidth]{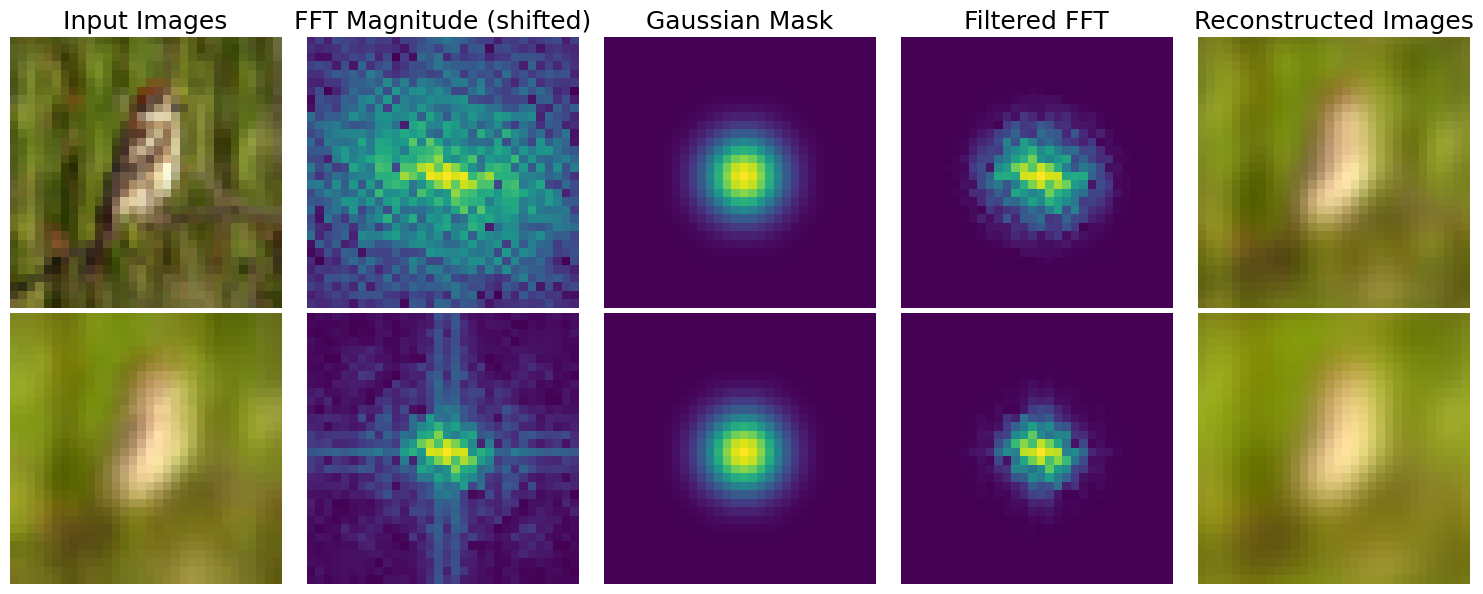}
    \caption{Visualization of \textit{FrequencyFilterCNN} processing on clean and corrupted images. The figure illustrates the key processing steps: original input images (first column), Gaussian mask applied in frequency domain (second column), FFT magnitude spectrum (third column), filtered frequency spectrum (fourth column), and resulting reconstructed images (fifth column). The top row shows processing of a clean bird image, while the bottom row demonstrates the same processing applied to a bird image corrupted with defocus blur at severity level 1.}
    \label{fig:robust_conv_visualization}
\end{figure}

\section{Results and Analysis}\label{sec:results}
\subsection{Overall Performance}
Table \ref{corruption_accuracy_table} compares the performance of our proposed methods against the baseline ResNet-18 (regular CNN layers)  on both clean CIFAR-10 test data and corrupted CIFAR-10-C data. Both of our proposed methods show improved robustness on CIFAR-10-C compared to the baseline. Method 2 (Supervised Contrastive Learning) achieves the best performance on both clean and corrupted data, with a 0.8\% improvement in mean corruption robustness accuracy over the baseline. Method 1 (Low-Frequency Activation Learning) shows a slight decrease in clean accuracy (-0.6\%) but with improvement in robustness accuracy (+0.3\%).

Supervised Contrastive Learning demonstrates balanced improvements across different corruption types, although it's performance is slightly lower than the baseline on noise corruptions (-0.40\% on average across all noises). To add an alternative perspective, we also present the results from Push-Pull CNN, that regularizes through architectural changes in CNN layer. It shows an exceptional performance specifically on noise corruptions (+13.97\% on average ). But on closer inspection, our methods still demonstrate a rather balanced performance improvement over a greater number of corruptions. These results suggest that the two methods have complementary strengths, with frequency-transform based approach particularly effective against blur corruptions, and contrastive-loss approach being performant against more numbers of individual corruptions.

\subsection{Analysis of the Methods}

\subsubsection{Low-Frequency Reconstruction}
\begin{figure}[t]
    \centering
    \begin{tabular}{cc}
        \begin{tabular}{p{3cm}} 
            \footnotesize\small {a) \textit{Pixelate} corruption with increasing severity} \\
        \end{tabular} &
        \includegraphics[width=0.7\textwidth,valign=m]{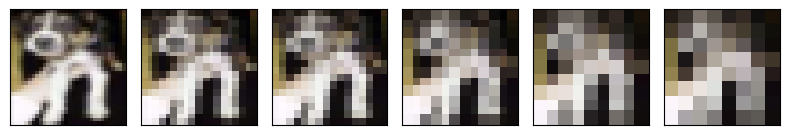}
        \\ 
        \begin{tabular}{p{3cm}} 
                \footnotesize\small {b) \textit{Regular CNN} activations} \\
        \end{tabular} &
        \includegraphics[width=0.7\textwidth,valign=m]{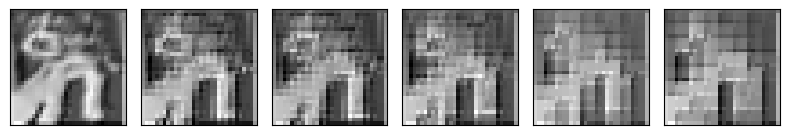}
        \\ 
        \begin{tabular}{p{3cm}} 
            \footnotesize\small {c) \textit{FrequencyFilterCNN}activations}
        \end{tabular} &
        \includegraphics[width=0.7\textwidth,valign=m]{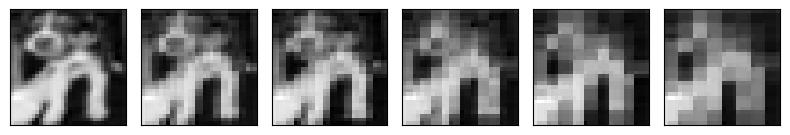} 
    \end{tabular}
    \caption{Comparison of activation patterns under pixelate corruption. (a) Original CIFAR-10 dog image (leftmost) and the same image corrupted with pixelate noise at increasing severity levels 1-5 (left to right). (b) Feature activations from a regular convolutional layer when processing the images shown in (a). (c) Feature activations from the proposed \textit{FrequencyFilterCNN} layer on the same input images. }
    \label{fig:pixelate-activation-comparison}
\end{figure}

Gaussian filtering at Frequency-domain reconstructs the original signal from low-frequency components only. From 
Fig.~\ref{fig:robust_conv_visualization}, we can see that, applying gaussian-mask on the Frequency Domain in input image, actually has effect of Blurring. 
 By training the network to produce similar activations for both original and their blurred counterpart, CNN learns robust kernels that are less prone to be activated by high-frequency patterns. Pixelate is another corruption where this approach benefits substantially. From Fig.~\ref{fig:pixelate-activation-comparison}, we find \textit{FrequencyFilterCNN} layer to demonstrate more robust activation patterns that are less sensitive to pixelate corruption compared to the respone maps of regular convolutional layer. We can see clearer feature representations maintained even at higher corruption severity.
The slight decrease in performance on noise corruptions (-1.91\%) suggests that this method may overly suppress high-frequency details that could be useful for distinguishing between certain classes under noise conditions. However, the overall corruption accuracy improves by 0.27\% compared to the baseline, indicating that the benefits outweigh this limitation.

\subsubsection{Supervised Contrastive Learning}
The regularization of cross-entropy loss with a supervised contrastive learning encourages the network to learn ``patterns that statistically prominent within different instances of a class''. For example, while cats may differ in breed and color, their facial structures tend to be similar — a property that distinguishes them clearly from other classes like dogs or inanimate objects such as tables. We asssume such features to be more shape-based than texture-based with shape being considered a more reliable cue for object identity. However, there is also the possibility that the network exploits unintended correlations to minimize the loss. For example, it might latch onto background elements or dataset-specific artifacts that co-occur frequently within certain classes, rather than the truly semantic object features. Our current work does not explicitly investigate the extent to which such shortcuts are being exploited.

\section{Conclusion and Future Work}\label{sec:conclusion}
In this paper, we explored two complementary regularization approaches to enhance CNN robustness that we hypothesize should discourage texture bias while promoting shape bias: a custom CNN architecture that enforces similarity between activations from original and low-frequency reconstructed input, and a supervised contrastive learning approach that should promote shape-focused representations.
Our results demonstrate that both methods effectively improve robustness against common corruptions on the CIFAR-10-C benchmark, with the supervised contrastive learning approach showing the strongest generic performance. 
In future work, we will evaluate our models on stylized datasets to quantify shape and texture bias more directly. We also plan to combine both regularization methods to exploit their complementary strengths. 

\end{document}